\documentclass[conference]{IEEEtran}
\IEEEoverridecommandlockouts
\usepackage{cite}
\usepackage{amsmath,amssymb,amsfonts}
\usepackage{algorithmic}
\usepackage{algorithm}
\usepackage{graphicx}
\usepackage{textcomp}
\usepackage{xcolor}
\usepackage{subcaption}
\usepackage{url}

\def\BibTeX{{\rm B\kern-.05em{\sc i\kern-.025em b}\kern-.08em
    T\kern-.1667em\lower.7ex\hbox{E}\kern-.125emX}}
\begin{document}

\title{
Robust Anomaly Detection with Graph Neural Networks using Controllability \\
}

\author{\IEEEauthorblockN{Yifan Wei}
\IEEEauthorblockA{\textit{Department of Computer Science}\\
Vanderbilt University \\
Nashville, TN, USA \\
yifan.wei@vanderbilt.edu}
\and
\IEEEauthorblockN{Anwar Said}
\IEEEauthorblockA{\textit{Department of Computer Science}\\
Vanderbilt University \\
Nashville, TN, USA \\
anwar.said@vanderbilt.edu}
\and
\IEEEauthorblockN{Waseem Abbas}
\IEEEauthorblockA{\textit{Department of Systems Engineering}\\
University of Texas at Dallas \\
Richardson, TX, USA \\
waseem.abbas@utdallas.edu}
\and
\IEEEauthorblockN{\centerline{Xenofon Koutsoukos}
\IEEEauthorblockA{
\textit{Department of Computer Science}\\
Vanderbilt University \\
Nashville, TN, USA \\
xenofon.koutsoukos@vanderbilt.edu}}
}

\maketitle

\begin{abstract}
Anomaly detection in complex domains poses significant challenges due to the need for extensive labeled data and the inherently imbalanced nature of anomalous versus benign samples. Graph-based machine learning models have emerged as a promising solution that combines attribute and relational data to uncover intricate patterns. However, the scarcity of anomalous data exacerbates the challenge, which requires innovative strategies to enhance model learning with limited information. In this paper, we hypothesize that the incorporation of the influence of the nodes, quantified through average controllability, can significantly improve the performance of anomaly detection. We propose two novel approaches to integrate average controllability into graph-based frameworks: (1) using average controllability as an edge weight and (2) encoding it as a one-hot edge attribute vector. Through rigorous evaluation on real-world and synthetic networks with six state-of-the-art baselines, our proposed methods demonstrate improved performance in identifying anomalies, highlighting the critical role of controllability measures in enhancing the performance of graph machine learning models. This work underscores the potential of integrating average controllability as additional metrics to address the challenges of anomaly detection in sparse and imbalanced datasets.
\end{abstract}


\section{Introduction}


Network Control Theory (NCT) offers a mathematical framework for analyzing and influencing the behavior of dynamic systems \cite{b1}. Within this framework, average controllability serves as a pivotal measure, evaluating how effectively a network's state can be transitioned from an initial to a desired final state through targeted control of specific nodes \cite{b2}. This measure extends its utility by quantifying the capacity of individual nodes to influence the network's overall dynamics, providing a distinctive perspective on their importance and role. Nodes with a higher average controllability exhibit greater influence on the behavior of the network, making them central to the structure and function of the network \cite{b3}. Using average controllability, one can identify nodes that are critical for guiding network behavior and gain deeper insights into the graph topology. These insights are particularly valuable in applications such as graph anomaly detection, where the influence of key nodes can help uncover irregular patterns and anomalies that conventional methods might overlook \cite{b24}.

Graph Anomaly Detection (GAD) is a vital area of study, focusing on identifying unusual patterns or outliers in graph-structured data. These anomalies can manifest as irregularities in the structure or attributes of nodes and edges, which may indicate significant, often hidden, events such as fraud \cite{b23}, and network intrusions \cite{b22}. Traditional Machine Learning (ML) methods for anomaly detection struggle to capture the complex nature of anomalies due to the limited availability of labeled data and many other known issues, making it challenging to accurately model the behavior of anomalous samples \cite{b25}. 

Graph Neural Networks (GNNs) have emerged as powerful tools in this context, using graph structure to enhance detection capabilities by learning both local and global patterns inherent in the data \cite{b4}. GNNs utilize both node attributes and relational information between nodes, allowing them to capture complex dependencies that go beyond individual features. By integrating node features with their neighbors’ information, GNNs can iteratively aggregate and update node representations through message passing layers. This aggregation scheme, often involving functions like mean, sum, or max pooling, ensures that each node’s final representation reflects not only its own characteristics but also the structure of its surrounding neighborhood \cite{b5}. 

Despite the strengths of GNNs, they also face notable limitations. The message passing mechanisms, whether using attention-based models or standard aggregation techniques, often fail to capture the unique behaviors of anomalous nodes. Anomalous nodes may exhibit patterns that deviate significantly from their neighbors, and conventional aggregation schemes tend to smooth out these distinct features by focusing on the average characteristics of the neighborhood \cite{b26}. As a result, these models may overlook or misrepresent critical anomalies.

To address these limitations, leveraging concepts from control theory—particularly average controllability—provides a more robust framework. Average controllability quantifies the importance of each node based on its capacity to control or influence the system's behavior \cite{b1, b2}. By incorporating this measure, we can identify nodes with disproportionate control or influence and integrate this information into the message passing process. This approach enhances the detection of key indicators of anomalous behavior in complex networks. Mainly, our contributions are as follows.

\begin{itemize}

\item \textbf{Unique Representation of Average Controllability:} We introduce a novel method to encode average controllability into GNNs through two distinct strategies. First, we represent it as an \emph{edge weight}, directly assigning weights to reflect the control influence of each connection. Second, we encode it as an \emph{edge attribute} using rank-based encoding \cite{b3}, capturing the relative importance of edges in terms of their control capacity in an attributed form.

\item \textbf{Evaluation and Comparison:} Extensive experiments were conducted using six benchmark GNNs across five anomaly detection datasets. The results demonstrate that our proposed approach effectively enhances anomaly detection performance.

\end{itemize}

By integrating these innovative metrics and exploring novel graph representations, this work aims to advance the capability of GNNs in detecting anomalies within graphs.

\section{Related Work}

Recent advances in graph anomaly detection have introduced several innovative methods. Class Label-aware Graph Anomaly Detection (CLAD) \cite{b28} leverages limited labeled nodes to enhance detection by incorporating class label information, improving structural anomaly identification. Graph Anomaly Detection via Neighborhood Reconstruction (GAD-NR) \cite{b29} is a recent variation of Graph Auto-Encoders (GAEs) that leverages the local structure, self-attributes, neighbor attributes and node representation reconstructs the neighborhood of nodes to distinguish between normal and anomalous nodes using reconstruction loss. Anomaly-Denoised Autoencoders (ADA-GAD) \cite{b30} employ anomaly-denoised augmentation to pretrain graph autoencoders at multiple levels, and introduce anomaly distribution regularization to mitigate overfitting issues. Hierarchical Memory Networks (HimNet) \cite{b31} learn hierarchical memory modules at node and graph levels via a GAE architecture and utilize node-level memory module for local anomaly detection and graph-level for holistic anomaly detection. Multi-representation Space Separation \cite{b32} for Graph-level Anomaly Detection designs an anomaly-aware module to separate normal and abnormal representations at node and graph levels. PREM \cite{b33} focuses on node-level anomaly detection using GNNs to capture complex patterns and improve detection accuracy in intricate graph structures.

Controllability, a fundamental concept in control theory, has also gained significant attention in graph representation learning and graph anomaly detection lately. Recent works leverage the controllability properties of dynamical networks to develop more expressive and computationally efficient graph representations for downstream anomaly detection tasks. One notable approach employs control properties to create augmented graph structures for contrastive learning. By perturbing the graph while preserving its controllability, this method generates augmented graphs that retain key structural characteristics, leading to superior performance in downstream classification tasks. Another line of research treats graphs as networked dynamical systems, where controllability is used to analyze the relationship between the graph’s topology and its control behavior \cite{b21}. By utilizing metrics such as the controllability Gramian, new graph representations are designed to capture both local and global structural information. 

\section{Methodology}


\begin{figure*}[!htb]
    \centering
    \includegraphics[width=0.99\linewidth]{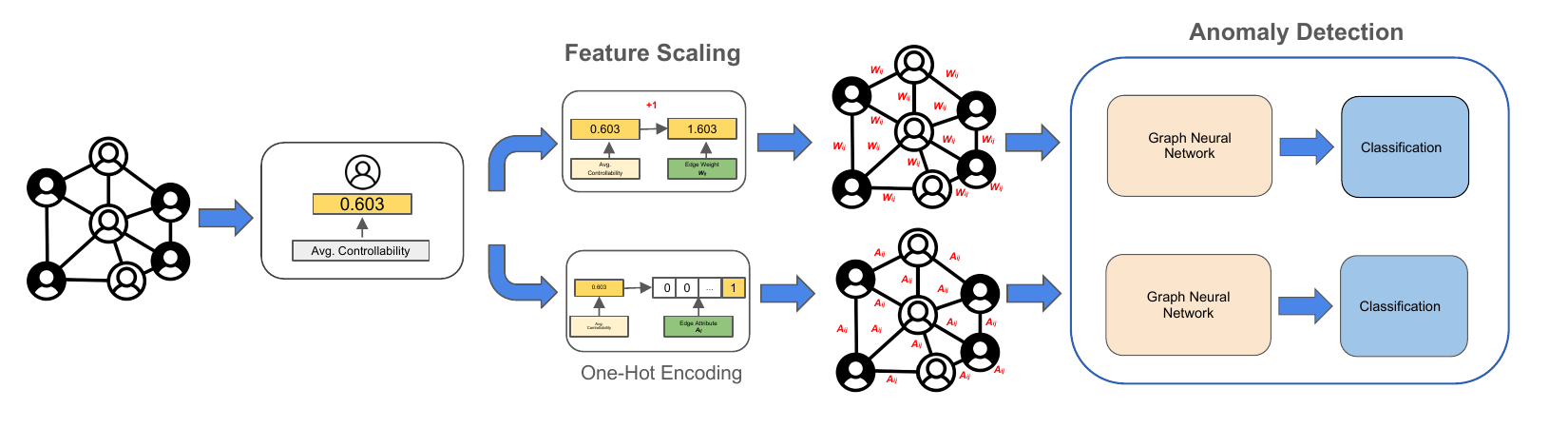}
    \vspace{-0.2in} 
    \caption{Illustration of the proposed methodology: The process begins with the computation of average controllability from the input network, which is subsequently integrated into the network topology. Following this, GNNs are trained to carry out anomaly detection.}
    \label{fig:illustration}
    \vspace{-0.2in}
\end{figure*}

Our proposed approach for GAD consists of three key steps: leveraging the NCT metric, average controllability, to quantify the influence of each node; constructing a network using derived parameters based on these metrics; and training a GNN model on the constructed graph for anomaly detection. In this section, we begin by introducing the necessary notations and providing a detailed overview of NCT and average controllability. Subsequently, we describe the process of network construction, followed by the methodology for training GNNs to effectively detect anomalies in the graph. We illustrate our proposed approach in Figure \ref{fig:illustration}.

\subsection{Network Control Theory}


Network Control Theory is a mathematical framework designed to analyze and influence the behavior of complex dynamic systems \cite{b2}. By identifying key control nodes within a graph—akin to influential nodes in a network—NCT enables the refinement of graph representations to capture not only the structural characteristics of these nodes but also their capacity to influence overall graph behavior. This includes their ability to propagate information efficiently and reconfigure relationships within the graph to achieve specific objectives \cite{b21}.

The structural foundation of NCT is defined by a network represented by the adjacency matrix \( A \in \mathbb{R}^{N \times N} \), where \( N = |V| \) denotes the number of nodes, and a control set matrix \( B \in \mathbb{R}^{N \times m} \). In this context, the temporal dynamics of each node, represented as \( x_i(t) \), are driven by a composite function that reflects the combined influence of all other nodes, \( x_j(t) \), along with external inputs \( u(t) \), modulated by weights in the network's topology. The evolution of node states can be described in terms of rates of change, where prior node activities impact the rate at which subsequent nodes' states evolve. This interaction is modeled by a differential equation:

\begin{equation}
    \label{eq:control}
    \frac{d}{dt}\boldsymbol{x}(t) = A\boldsymbol{x}+B\boldsymbol{u}(t)
\end{equation}

where \( \boldsymbol{x}(t) = [x_1(t), x_2(t),\ldots, x_n(t)]^\top \) is the vector of node states, \( A \) is the adjacency matrix, and \( \boldsymbol{u}(t) = [u_1(t), u_2(t),\ldots, u_n(t)]^\top \) is the vector of control signals. The matrix \( B \in \mathbb{R}^{N \times m} \) quantifies the influence of the inputs on each node. In our experiments, \( B \in \mathbb{R}^{N \times N} \) is set as the identity matrix, indicating a fully controllable system.

A key mathematical tool in network control is the controllability Gramian, which measures the ease with which the system can be driven from one state to another using control inputs \cite{b3}. In the system defined by the equation above, the infinite horizon controllability Gramian is given by:

\begin{equation}
    \label{eq:Gramian}
    \mathcal{W} = \int_{0}^{\infty}e^{-A\tau}(-B)(-B)^\top e^{-A^\top \tau}d\tau \; \in \; \mathbb{R}^{N_f \times N_f}.
\end{equation}

For stable systems—where all eigenvalues of \( -A \) have negative real parts—the Gramian \( \mathcal{W} \) converges and can be derived from the Lyapunov equation \cite{b3}:

\begin{equation}
    \label{eq:Lyapunov}
    (-A)\mathcal{W} + \mathcal{W}(-A)^\top + (-B)(-B)^\top  = 0,
\end{equation}

The system described by \eqref{eq:control} allows us to compute several NCT metrics, with the average controllability metric being a primary focus of this study.

\subsection{Average controllability}

Average controllability measures the ability of an individual node to influence the overall dynamics of a network \cite{b2}. Nodes with varying degrees of average controllability exhibit different levels of impact on the network's behavior, highlighting their unique roles within the graph structure. Building on this insight, our study focuses on transforming average controllability values into graph-related features such as \emph{edge attributes} and \emph{edge weights}. By embedding these features into the graph, we enrich the original graph data with additional topological information, aiming to improve the accuracy of GAD. The process begins with calculating the average controllability for each node, a step detailed in Algorithm \ref{AC-Algo}, which serves as the foundation for constructing enhanced graph representations tailored for GNN-based anomaly detection.

\begin{algorithm}
\caption{Compute Average Controllability Score}
\label{AC-Algo}
\begin{algorithmic}[1]
\STATE \textbf{Input:} $G = (V, E)$
\STATE \textbf{Output:} Average Controllability (AC) score for individual nodes
\STATE $A \gets$ Dense adjacency matrix from $G$
\STATE $B \gets \mathbf{I}_{|V|}$ \emph{(Identity matrix of size $|V|$)}
\STATE \textbf{Compute Eigenvalues:} \\
\hspace{2em} Solve the characteristic equation: 
\[
\det(A - \lambda \mathbf{I}) = 0
\]
\STATE $\{\lambda_1, \lambda_2, \dots, \lambda_{|V|}\} \gets$ Eigenvalues of $A$
\STATE $l \gets \max \left( |\lambda_1|, |\lambda_2|, \dots, |\lambda_{|V|}| \right)$
\STATE $A_{\text{norm}} \gets \frac{A}{l + 1} - \mathbf{I}_{|V|}$ \emph{(Normalize $A$)}
\STATE $t = [t_0, t_1, t_2, \dots, t_n]$ \emph{(Time steps for integration)}
\STATE $dE \gets e^{A_{\text{norm}} \cdot \Delta t}$ \emph{(State transition matrix for time step $\Delta t$)}
\STATE $dEa \gets \mathbf{I}_{|V|}$ \emph{(Initialize state transition matrices)}
\STATE $dG \gets 0$ \emph{(Initialize Gramian matrix)}

\FOR{each $i$ in $t$}
    \STATE $dEa_{t_i} \gets dEa_{t_{i-1}} \cdot dE$ \emph{(Update state transition matrix)}
    \STATE $dEab \gets dEa_{t_i} \cdot B$ \emph{(Control matrix contribution at $t_i$)}
    \STATE $dG \gets dG + (dEab \cdot dEab^\top) \cdot \Delta t$ \emph{(Update Gramian matrix)}
\ENDFOR

\STATE $\mathcal{W} \gets dG$ \emph{(Final controllability Gramian)}
\STATE $AC \gets \text{diag}(\mathcal{W})$ \emph{(Extract diagonal entries as node scores)}
\RETURN $AC$
\end{algorithmic}
\end{algorithm}

The algorithm \ref{AC-Algo} starts by taking the input graph \( G \) and extracting its dense adjacency matrix \( A \). In step 6-7, it computes the largest absolute eigenvalue \( l \) of \( A \). In step 8, the matrix \( A \) is normalized by dividing it by \( l+1 \) and subtracting the identity matrix. Step 9 defines a series of time steps, and in step 10, the matrix exponential \( e^{A_{\text{norm}} \cdot \Delta t} \) is computed to model how the system evolves over time. Steps 11 and 12 initialize the state transition and controllability Gramian matrices. In steps 13-17, the algorithm iterates through each time step, updating the state transition matrix and computing the controllability Gramian by multiplying the state transition matrix with the control matrix and its transpose. Finally, the Gramian is integrated over time, and the diagonal entries of the resulting matrix give the average controllability score for each node, which is returned as the output. 

\subsection{Network construction with derived parameters}

We utilize average controllability to inform the construction of two critical parameters for the GNN: the edge weight and edge attributes. The edge weight parameter typically defines the strength or importance of the connections between nodes, while edge attributes provide additional information to the message passing  mechanism about these connections. Average controllability influences both of these parameters, therefore our study embedded the information contained in average controllability into these two parameters.

\subsubsection{Edge Weights}  
Traditional GNNs typically operate on undirected graphs, where connections between nodes are inherently bidirectional. To maintain compatibility with this structure, we preprocess the graph by adding reciprocal edges, ensuring every connection is bidirectional. Additionally, we observed that the average controllability scores are constrained between 0 and 1, making it challenging to distinguish differences across nodes. To address this limitation, we enhance differentiation by adding 1 to each average controllability score. The edges are then assigned weights proportional to the average controllability of their source nodes, ensuring that edges originating from highly controllable nodes receive greater weight. This weighting scheme enables GNNs to focus on connections that are critical to the network's overall controllability, thereby enhancing the model's ability to detect anomalies that disrupt these pivotal connections.  

\subsubsection{Edge Attributes}  
As a second contribution, we propose encoding average controllability into a vector representation in the form of edge attributes, providing additional context to the message passing process. The procedure is as follows:  

Given the average controllability vector for all nodes in the graph, computed as described in Algorithm \ref{AC-Algo}, where each entry corresponds to a node, we follow the approach presented in \cite{b3} and construct a histogram \( \mathcal{H} \) with \( k \) bins to capture the distribution of controllability values. The range of \( \mathcal{H} \) spans from the minimum to the maximum average controllability value, with equal-width bins covering the entire range. Each bin represents a specific range of controllability values, and the height of each bin indicates the frequency of nodes whose controllability falls within that range. This histogram summarizes the distribution of controllability across nodes, highlighting the prevalence of specific controllability levels.  

To generate a feature vector for each edge \( e = (v_s, v_t) \) using \( \mathcal{H} \), we employ a one-hot encoding strategy based on the average controllability of the source node \( v_s \). The one-hot encoded feature vector for edge \( e \), at index \( i \), is denoted as \( \mathbf{h}^0_e(i) \), where the index corresponds to the bin in \( \mathcal{H} \) that contains the average controllability value of \( v_s \).

\begin{equation}
\label{eq:rank-encoding}
\mathbf{h}^0_e(i) = 
\begin{cases}
1 & \text{if } C_a(v_s) \in \mathcal{H}(i) \\
0 & \text{otherwise}
\end{cases}
\end{equation}
We present the algorithm for constructing edge attribute in Algorithm \ref{algo:acencoding}.

\begin{algorithm}[!t]
\caption{Average Controllability Encoding}
\label{Rank Encoding ALgo}
\begin{algorithmic}[1]
    \STATE \textbf{Input:} Graph $G = (V, E)$, number of bins $k$ 
    \STATE \textbf{Output:} Edge Encoding matrix $E_x$
    \STATE $AC \gets$ Average Controllability($G$)
    \STATE $\mathcal{H} \gets$ Construct histogram $\mathcal{H}$ with $k$ number of bins
    \STATE $E_x \gets$ initialize edge feature matrix
    \FOR{each edge $e$ in $E$}
        \STATE $v_s \gets$ get the source node
        \STATE $a_c \gets AC[v_s]$: 
        \STATE $index \gets $ retrieve the corresponding bin index of $a_c$ from $\mathcal{H}$ using equation \ref{eq:rank-encoding}
        \STATE $e_x \gets $: initialize zero vector of size $k$
        \STATE $e_x[index - 1] \gets 1$
        \STATE $E_x[v_s] \gets e_x$
    \ENDFOR 
    
    \STATE \textbf{Return} $E_x$
\end{algorithmic}
\label{algo:acencoding}
\end{algorithm}


\subsection{Graph Neural Networks}

Graph Neural Networks (GNNs) are a class of neural networks designed to operate on graph-structured data \cite{b4}. Unlike traditional neural networks that process feature vectors, GNNs process node, edge, and graph-level information to extract patterns and perform predictions. The fundamental mechanism that drives GNNs is message passing, where nodes iteratively exchange information with their neighbors and aggregate that information to update their own representations \cite{b27}. By passing messages through the graph, GNNs can better capture both the structural information of the graph and the features associated with each node. 

The message passing process at each layer \( l \) consists of two main steps: message aggregation and node update \cite{b27}. 

\subsubsection{Message Passing}
The general form of message aggregation is defined as:
\begin{equation}
  m_i^{(l)} = \sum_{j \in \mathcal{N}(i)} M(h_i^{(l)}, h_j^{(l)}, e_{ij})
  \label{eq:message_aggregation_mpnn}
\end{equation}
Where \( m_i^{(l)} \) is the message for node \( i \) at layer \( l \), \( \mathcal{N}(i) \) denotes the set of neighboring nodes of node \( i \), \( h_i^{(l)} \) and \( h_j^{(l)} \) are the hidden states of the node \( i \) and its neighbor \( j \) at layer \( l \), \( e_{ij} \) represents edge features between nodes \( i \) and \( j \), and \( M(\cdot) \) is the message function.

\subsubsection{Node Update}
After aggregating messages from its neighbors, the node's hidden state is updated using the following equation:
\begin{equation}
  h_i^{(l+1)} = U(h_i^{(l)}, m_i^{(l)})
  \label{eq:node_update_mpnn}
\end{equation}
Where \( h_i^{(l+1)} \) is the updated hidden state of node \( i \) at layer \( l+1 \), \( h_i^{(l)} \) is the current state of node \( i \) at layer \( l \), \( m_i^{(l)} \) is the aggregated message from neighboring nodes at layer \( l \), and \( U(\cdot) \) is the update function.


We propose incorporating \emph{edge\_weight} and \emph{edge\_attributes} as \( e_{ij} \) to enhance the message passing process in GNNs. 

\emph{Edge\_weight} \( w_{ij} \) modulate the influence of neighbor nodes on the central node. By incorporating edge weights, the message passing process becomes more fine-grained, allowing the model to prioritize information from more influential neighbors. The modified message aggregation step becomes: 
\begin{equation}
   m_i^{(l)} = \sum_{j \in \mathcal{N}(i)} w_{ij} \cdot M(h_i^{(l)}, h_j^{(l)})
\end{equation}
where edge weights \( w_{ij} \) scale the contribution of each neighbor \( j \) to node \( i \)'s update, which helps the neural network to aggregate edge information with importance. 

\emph{Edge\_attribute} \( e_{ij} \) provide an additional measure to aid the neural network in understanding the discrete relationship between nodes. Instead of treating all edges as homogeneous, edge attributes, like one-hot encoded vectors, allow the GNNs to incorporate relationship-specific details, leading to more informative message passing. 

Once the average controllability information is injected to the network structure, we use state-of-the-art GNN approaches to evaluate the proposed approach. We discuss those details in the next section.

\section{Numerical Evaluation}

\subsection{Experimental Setup}

\subsubsection{Baselines}
To evaluate the effectiveness of the proposed approaches, we chose a diverse set of graph convolutional models based on their compatibility with the introduced features. Six models including $k-$GNN \cite{b4}, BGNN \cite{b6}, SGC \cite{b7}, GIN \cite{b8}, GraphSAGE \cite{b5}, and TAG \cite{b9}—utilize edge weights during their forward pass, leveraging the weighted connections to enhance their performance. Additionally, four models—GEN \cite{b10}, ResGatedGraph \cite{b11}, GAT2 \cite{b12}, and UniMP \cite{b13}—incorporate edge attributes, allowing them to benefit from the added contextual information provided by these features. A brief overview of each model is provided below.

\textbf{Baseline GNN Convolutions with Edge Weight} \\
\textbf{$k-$GNN:} It leverages a localized, first-order approximation of spectral graph convolutions to efficiently aggregate features from neighboring nodes. This approach enables GCNs to encode both local graph structure and node features into low-dimensional embeddings, which can be used for tasks such as node classification and link prediction \cite{b4}. 
\textbf{BGNN:} Combines GNNs with gradient boosting decision trees (GBDT), preprocessing node features for both models to iteratively enhance predictions \cite{b6}.
\textbf{SGC}: A simplified GCN that reduces the complexity of graph convolution by removing non-linearities and collapsing weight matrices across layers \cite{b7}.
\textbf{GIN:} A message passing framework model designed to capture graph isomorphism, enhancing feature aggregation through GINConv \cite{b8}.
\textbf{GraphSAGE:} Uses inductive learning by sampling and aggregating node features from local neighborhoods \cite{b5}.
\textbf{TAG:} Employs fixed-size learnable filters to perform convolution, enhancing efficiency with a localized approach \cite{b9}. 

\textbf{Baseline Models Supporting Edge Attributes}\\

\textbf{GEN:} Introduces energy dynamics into graph convolution, emphasizing influential nodes and edges to highlight structural patterns important for tasks like anomaly detection \cite{b10}. \textbf{ResGatedGraph}, uses residual connections to enable deeper networks by retaining original node information, helping to mitigate the vanishing gradient problem \cite{b11}.
\textbf{GATv2d}, enhances attention mechanisms from the original GAT, improving the precision of neighbor selection and feature aggregation with multiple attention heads \cite{b12}.
\textbf{UniMP}, adapts the self-attention mechanism from sequence models to graph data, allowing nodes to attend to all others in the graph, enabling the capture of long-range dependencies and global interactions \cite{b13}.

\subsubsection{Datasets}

 We utilize three real-world anomaly detection datasets including Reddit \cite{b14}, Amazon \cite{b14} and FraudAmazon \cite{b7}. The reddit dataset was obtained from the GAD Benchmark consisting of \( 10,984 \) nodes, of which \( 3.3\% \) are labeled as anomalies, and \( 168,016 \) edges. Each node is represented by \( 64 \) features, derived from text embeddings. Relationships are defined based on nodes appearing in the same post. The Amazon dataset was also obtained from the GAD Benchmark \cite{b14}, and the FraudAmazon dataset from DGL \cite{b7}. The FraudAmazon dataset includes product reviews in the Musical Instruments category. Users with more than \( 80\% \) helpful votes are labeled as benign entities, while those with fewer than \( 20\% \) helpful votes are labeled as fraudulent entities. Each node is described by \( 25 \) handcrafted features. This dataset consists of \( 11,944 \) nodes and three types of edge allocations: \( U \text{-} P \text{-} U \), \( U \text{-} S \text{-} U \), and \( U \text{-} V \text{-} U \).  For the synthetic setting, due to the lack of publicly available datasets that explicitly define structural and contextual anomalies, we modify well-known node classification dataset, Cora \cite{b15} by injecting anomalies. Specifically, we introduce both structural and contextual anomalies to evaluate the proposed methods.  

For structural anomalies, we first select \( m \times n \) nodes to form structural anomaly clusters. These nodes are divided into \( n \) groups, each containing \( m \) nodes. Within each group, all possible pairs of nodes are considered for potential new edges. A new edge is added between a pair of nodes with a probability of \( 1 - p \), provided the edge does not already exist in the graph. This process modifies the local connectivity, rendering these groups structurally anomalous. The selected nodes are labeled as anomalies with a value of \( 1 \) in the label tensor.  

For contextual anomalies, we randomly select \( m \times n \) nodes, denoted as \( V_c \), to serve as contextual anomalies. The remaining nodes are stored in \( V_r \). For each node in \( V_c \), we randomly select \( q \) nodes from \( V_r \). Among these \( q \) nodes, the one with the most dissimilar feature vector (measured by Euclidean distance) is identified. The feature vector of this dissimilar node is then copied to the corresponding node in \( V_c \), making its features significantly different from those of its neighboring nodes. The nodes in \( V_c \) are labeled with a value of \( 1 \) in the label tensor, indicating contextual anomalies. All other nodes in the dataset are considered normal.

\begin{table*}[!t]
\caption{Reddit dataset: Performance comparison of the proposed approach with GCN, BGNN, SGC, GIN, GraphSAGE, and TAG as graph convolution methods.}
\begin{center}
\begin{tabular}{|l|c|c|c|c|c|c|c|c|c|c|c|c|}
\hline
\textbf{Metrics} & \multicolumn{2}{|c|}{\textbf{GCN}} & \multicolumn{2}{|c|}{\textbf{BGNN}} & \multicolumn{2}{|c|}{\textbf{SGC}} & \multicolumn{2}{|c|}{\textbf{GIN}} & \multicolumn{2}{|c|}{\textbf{GraphSAGE}} & \multicolumn{2}{|c|}{\textbf{TAG}} \\
\cline{2-13} 
& \textbf{\textit{Ours}} & \textbf{\textit{Baseline}} & \textbf{\textit{Ours}} & \textbf{\textit{Baseline}} & \textbf{\textit{Ours}} & \textbf{\textit{Baseline}} & \textbf{\textit{Ours}} & \textbf{\textit{Baseline}} & \textbf{\textit{Ours}} & \textbf{\textit{Baseline}} & \textbf{\textit{Ours}} & \textbf{\textit{Baseline}} \\
\hline
\textbf{AUROC} & \textbf{0.615} & 0.613 & \textbf{0.687} & 0.682 & 0.553 & 0.553 & 0.596 & \textbf{0.649} & 0.646 & \textbf{0.649} & \textbf{0.646} & 0.615 \\
\hline
\textbf{AUPRC} & \textbf{0.050} & 0.046 & \textbf{0.067} & 0.066 & 0.047 & 0.047 & 0.060 & \textbf{0.061} & \textbf{0.059} & 0.057 & \textbf{0.059} & 0.047 \\
\hline
\textbf{RecK} & \textbf{0.068} & 0.061 & \textbf{0.068} & 0.054 & 0.082 & 0.082 & \textbf{0.109} & 0.075 & \textbf{0.088} & 0.061 & \textbf{0.088} & 0.054 \\
\hline
\end{tabular}
\label{tab:model_comparison_adjusted}
\end{center}
\end{table*}
\begin{figure*}[!t]
    \centering
    \begin{subfigure}{0.49\textwidth}
        \centering
        \scalebox{0.33}{
        \includegraphics{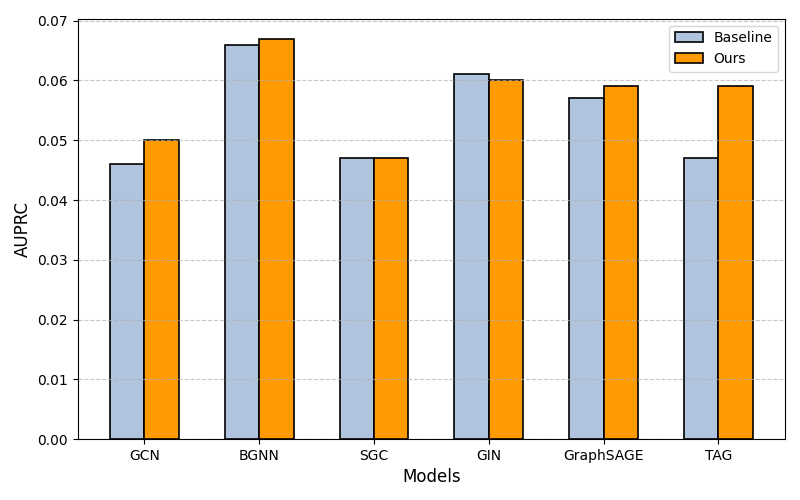}}
        \label{fig:auprc_histogram1}
    \end{subfigure}
    \hfill
    \begin{subfigure}{0.49\textwidth}
        \centering
        \scalebox{0.33}{
        \includegraphics{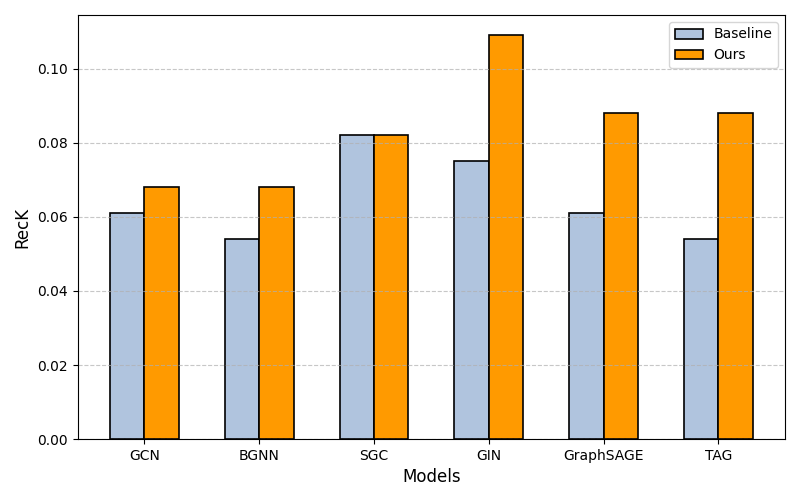}}
        \label{fig:rec_histogram1}
    \end{subfigure}
    \vspace{-0.15in} 
    \caption{Comparative analysis of performance metrics AUPRC (left) and RecK (right) via edge weight a achieved by different models on the Reddit Dataset.}
    \label{fig:combined_figures}
    \vspace{-0.2in}
\end{figure*}

\subsubsection{Evaluation Metrics}

We use Area Under the Receiver Operating Characteristic Curve (AUROC), the Area Under the Precision-Recall Curve (AUPRC) calculated using average precision, and the Recall score within the top-\( K \) predictions (\( \text{Rec@K} \)) as performance metrics for the GAD task \cite{b14}. The parameter \( K \) is set to the number of anomalies within the test set. For all metrics, anomalies are treated as the positive class, with higher scores indicating better model performance.  

Each metric serves a specific purpose: AUROC evaluates overall performance but is not sensitive to top-\( K \) predictions, \( \text{Rec@K} \) focuses exclusively on top-\( K \) performance, and AUPRC strikes a balance between the two. For example, suppose the test set contains \( 10 \) anomalies within \( 1000 \) data points, and a model ranks the anomalies in positions \( 11 \) to \( 20 \). In this case, the model would achieve an AUROC of \( 0.99 \), an AUPRC of \( 0.33 \), and a \( \text{Rec@10} \) of \( 0 \).  

We chose \( \text{Rec@K} \) over traditional recall because the order of predictions is critical in GAD. In practical applications, the top-ranked results are often prioritized for further analysis as potential anomalies. By assessing the proportion of true anomalies within the top-\( K \) predictions, \( \text{Rec@K} \) provides a more accurate measure of the model's ability to identify significant anomalies. This metric effectively evaluates how well the model prioritizes the most relevant anomalous nodes, which is crucial in scenarios where only the most suspicious cases merit further investigation.

\subsubsection{Parameters Setup} 

In this study, we utilize a step size of \(0.2 \) and treat the system as continuous to calculate the average controllability. Each GNN is constructed with two layers of convolutions and a multilayer perceptron (MLP) for classification. Specifically, the first GCN layer transforms the original node features into a \(32 \)-dimensional space, while the second GCN layer maintains the feature dimensions at \(32 \). The MLP layer performs the final binary classification. We apply the ReLU activation function throughout the network and set the dropout rate to \(0 \).

For models that incorporate edge attributes, we encode these attributes using a fully connected linear layer within the GNN context. The number of epochs for each trial is set to \(200 \), and the learning rate is fixed at \(0.01 \). To address the issue of class imbalance, we use a \emph{weighted cross-entropy loss function}, where the weight is defined as the ratio of benign nodes to anomalous nodes. The Adam optimizer is employed for model optimization. In the experiments involving edge attributes, bin sizes are introduced as a hyperparameter, and we evaluate each dataset with different bin sizes of \(5 \), \(20 \), \(30 \), and \(50 \). We ran each experiment \(10 \) times with \(10 \) different seeds and report the average scores.  
 
\subsection{Results with Edge Weights}



We evaluate the aforementioned baseline for GAD by comparing their performance with and without the use of the proposed approach, using AUROC, AUPRC, and RecK metrics on Reddit Dataset \cite{b14}. For a baseline, we train each models with the same hyper-parameters and architecture and then use the proposed approach to integrate the metrics and train the same model on the augmented data. We present the evaluation results in Table \ref{tab:model_comparison_adjusted}. We observe that there is a consistent boost to each model's performance using the proposed approach.


\begin{table*}[!htb]
\caption{Performance comparison of using average controllability as an edge attribute with GEN, RES, GAT2, and UNiMP}
\begin{center}
    \begin{tabular}[width = \textwidth]{|l|c|c|c|c|c|c|c|c|}
        \hline
        \textbf{Metrics} & \multicolumn{2}{|c|}{\textbf{GEN}} & \multicolumn{2}{|c|}{\textbf{RES}} & \multicolumn{2}{|c|}{\textbf{GAT2}} & \multicolumn{2}{|c|}{\textbf{UniMP}} \\
        \cline{2-9}
        & \textbf{Ours} & \textbf{Baseline} & \textbf{Ours} & \textbf{Baseline} & \textbf{Ours} & \textbf{Baseline} & \textbf{Ours} & \textbf{Baseline} \\
        \hline
        \multicolumn{9}{|l|}{\textbf{Reddit dataset}} \\
        \hline
        \textbf{AUPRC} & \textbf{0.085} & 0.070 & \textbf{0.061} & 0.046 & 0.056 & 0.056 & \textbf{0.071} & 0.062 \\
        \textbf{RecK} & \textbf{0.127} & 0.099 & \textbf{0.097} & 0.043 & 0.063 & 0.063 & \textbf{0.095} & 0.064 \\
        \hline
        \multicolumn{9}{|l|}{\textbf{Injected Cora dataset}} \\
        \hline
        \textbf{AUPRC} & \textbf{0.280} & 0.223 & \textbf{0.500} & 0.387 & 0.248 & 0.248 & \textbf{0.350} & 0.198 \\
        \textbf{RecK} & \textbf{0.310} & 0.303 & \textbf{0.483} & 0.386 & 0.441 & 0.441 & \textbf{0.455} & 0.324 \\
        \hline
        \multicolumn{9}{|l|}{\textbf{FraudAmazon dataset}} \\
        \hline
        \textbf{AUPRC} & 0.873 & \textbf{0.876} & \textbf{0.278} & 0.235 & 0.384 & 0.384 & \textbf{0.846} & 0.844 \\
        \textbf{RecK} & 0.834 & \textbf{0.839} & \textbf{0.345} & 0.281 & 0.411 & 0.411 & 0.803 & 0.803 \\
        \hline
    \end{tabular}
\end{center}
\label{tab:gcn_combined_comparison}
\end{table*}
\begin{figure*}[!t]
    \centering
    \begin{subfigure}[b]{0.9\textwidth}
        \centering
        \begin{subfigure}[b]{0.49\textwidth}
            \centering
            \scalebox{0.33}{
            \includegraphics{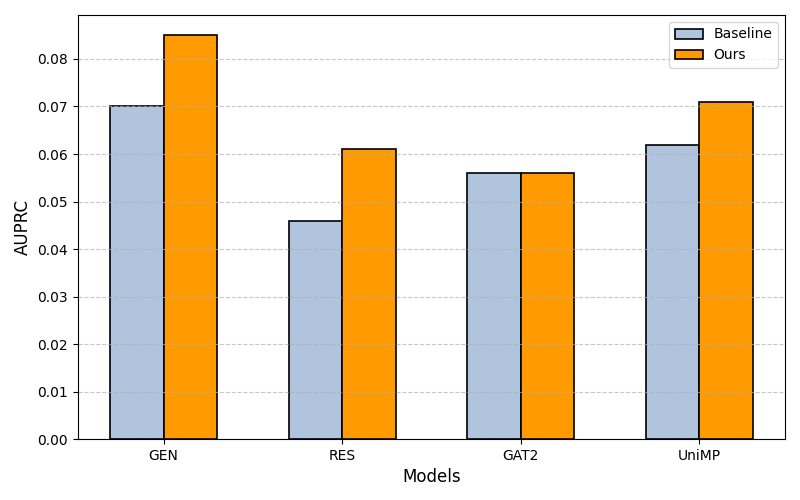}}
            \label{fig:auprc_histogram}
        \end{subfigure}
        \hfill
        \begin{subfigure}[b]{0.49\textwidth}
            \centering
            \scalebox{0.33}{
            \includegraphics{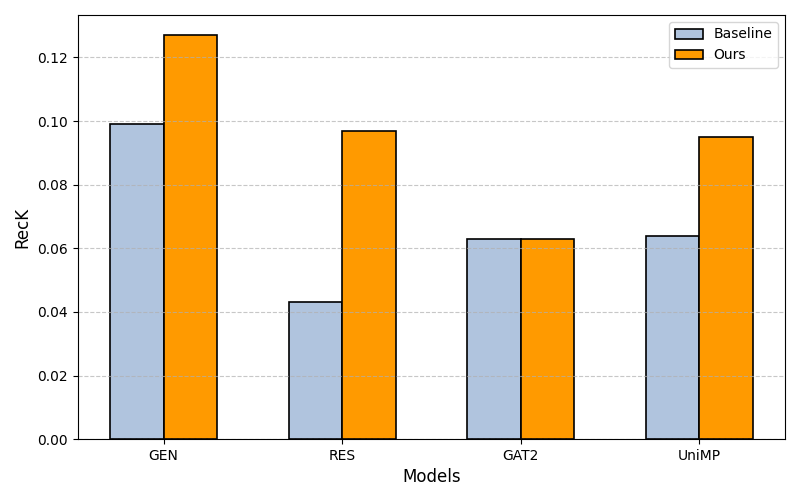}}
            \label{fig:rec_histogram}
        \end{subfigure}
    \end{subfigure}
    \vspace{-0.15in} 
    \caption{Comparison of performance metrics (AUPRC and RecK) for models on the Reddit Dataset with edge attribute approach.}
    \label{fig:combined_subfigure}
\end{figure*}
Specifically, the proposed approach achieves identical or improved performance across all models on the AUPRC metric, which balances precision and recall, and thereby, a better measure of the performance in this imbalanced Reddit dataset \cite{b14}. BGNN's AUPRC score increases from \(0.065\) to \(0.067\), indicating a balanced recognition of positive cases. GraphSAGE maintains a similar AUPRC score, highlighting the consistent predictive capability of the model in balancing precision and recall, even with different configurations. SGC and GIN models exhibit similar trends, reflecting the overall stability of the models' performance. The proposed approach als achieves a consistent increase across all models on the RecK metric, which evaluates how well models prioritize true anomalies in the top-K predictions. Notable improvements include a \(10.5\%\) increase for GCN. TAG's RecK shows a substantial increase of \(62.4\%\), underscoring its enhanced ability to prioritize true anomalies. Additionally, GraphSAGE's RecK increases by \(44.4\%\), and GIN's RecK shows a \(45.5\%\) increase. The result reflects its capacity to better rank positive cases. This consistent improvement underscores the essential role of the proposed approach in enhancing top-K ranking performance. For a clearer visualization of the results, we have illustrated the performance comparison in Figure \ref{fig:combined_figures}.

\subsection{Results with Edge Attributes}

For this task, we considered four models, both with and without the proposed approach and three datasets: Reddit, Fraud Amazon and Injected Cora dataset. For constructing the edge attributes, the optimal overall performance from configurations based on bin sizes of $10$, $20$, $30$, and $50$ is chosen. Our primary focus is on the AUPRC and RecK metrics. This focus is justified by two main factors: firstly, the significant imbalance present in each dataset, and secondly, the importance of an anomaly detection algorithm having a high probability of correctly identifying true anomalies. The numerical results are detailed in Table \ref{tab:gcn_combined_comparison} and also illustrated in Figure \ref{fig:combined_subfigure} for Reddit dataset. 

The Reddit and Injected Cora datasets show a significant increase in AUPRC scores for the GEN, RES, and UniMP models with the proposed approach, while the GAT2 model's performance remains unchanged. The baseline for the AUPRC score is determined by the fraction of positives, which corresponds to the percentage of anomalies in the dataset. For example, the Reddit dataset contains \(3.3\%\) anomalies, setting the baseline at \(0.033\). Although all models exceed this baseline in the baseline setting, the inclusion of edge attribute with the proposed approach still boosts their performance. Specifically, it improves the AUPRC for GEN by \(20.7\%\), RES by \(32.9\%\), and UniMP by \(14.6\%\). In the Injected Cora dataset, where \(50\) contextual and \(50\) structural anomalies are introduced, the baseline is set at \(3.7\%\) (\(0.037\)). Here, each model significantly exceeds the baseline, indicating that artificially injected anomalies may be easier to detect. The addition of edge attributes further enhances performance, with improvements of \(25.2\%\) for GEN, \(29.3\%\) for RES, and \(76.6\%\) for UniMP. Conversely, no such improvement is observed for the FraudAmazon dataset, which suggests that graph topology information may not be as crucial for anomaly detection in this dataset---a hypothesis that is supported by later experiments discussed in \ref{conclusion}.

Regarding metric RecK, which measures the algorithm's ability to recommend true anomalies effectively, the proposed approach also achieve progress. The Reddit and Injected Cora datasets exhibit a notable increase in successful anomaly recommendations across all models, except for GAT2. For the Reddit dataset, the percentage of successful recommendations increases from \(9.9\%\) to \(12.6\%\) for GEN, \(4\%\) to \(10\%\) for RES, and \(6.4\%\) to \(9.5\%\) for UniMP. Similarly, for the Injected Cora dataset, successful recommendations rise from \(30\%\) to \(31\%\) for GEN, \(38\%\) to \(48\%\) for RES, and \(32\%\) to \(46\%\) for UniMP. As with the AUPRC metric, no significant improvement is observed for the FraudAmazon dataset, except for a modest increase from \(28\%\) to \(34\%\) in the RES model.

\section{Conclusion and future works}
\label{conclusion}

In this paper, we introduce two novel approaches to incorporate average controllability into the network topology. The first approach utilizes edge weights to account for node influence within the message passing mechanism. The second approach employs an encoding scheme to transform average controllability into edge attributes. We evaluate these proposed methods against several baseline models and benchmark datasets. Our results indicate that incorporating edge attributes with average controllability significantly enhances model performance on datasets such as Reddit and Injected Cora, leading to substantial improvements in both AUPRC and RecK scores. 
Moreover, across all models and metrics, the inclusion of edge weights improves the models' ability to balance precision and recall, as evidenced by the gains in AUPRC, and to effectively recommend true anomalies, as reflected by RecK improvements. 

Future work will focus on investigating why the inclusion of edge attributes improved the predictive capabilities of GNNs on certain datasets, but had no noticeable impact on others. Our current hypothesis posits that edge attributes serve as an additional tool to enhance the structural information of the graph, making them beneficial only when this structural information is vital for distinguishing between anomalies and normal nodes. Overall, we believe this research will pave the way for future studies integrating network control theory and graph machine learning to develop more effective approaches for robust anomaly detection.

\section*{Acknowledgment}
This work is supported by the National Science Foundation under Grant Numbers 2325416 and 2325417.



\end{document}